\documentclass{article}
\usepackage{spconf,amsmath,amssymb,graphicx,booktabs,hyperref}
\usepackage{hyphenat,balance}
\hypersetup{hidelinks}
\newcommand{\R}{\mathbb{R}}
\newcommand{\pos}[1]{[#1]_+}
\title{Hexagonal Basis Projectors for Model-Based Iterative CT Reconstruction}
\name{Ke Chen, Alireza Entezari\thanks{\normalsize This research is supported in part by NSF grant CCF-2210866.}}
\address{University of Florida}
\begin{document}
\ninept
\maketitle
\begin{abstract}
Image representation affects both the forward model and quantitative feature recovery in model-based iterative CT reconstruction. For isotropically bandlimited signals, the hexagonal lattice is the most sampling-efficient lattice in two dimensions, with greater directional isotropy and a larger alias-free bandwidth than the Cartesian grid at equal density. We develop constant and linear hexagonal projectors for two-dimensional fan-beam CT using box spline calculus. Both bases form a partition of unity on common coefficient centers; the linear tent also reproduces affine functions. We derive analytic ray footprints and effective detector-blur expressions and implement efficient GPU forward and backprojection. On simulated FORBILD data, the linear model improves on Cartesian and hexagonal pixels, with its best noisy-data accuracy at the coarser resolution. Real-world 2DeteCT comparisons use high-dose, high-resolution Cartesian reconstructions as pseudo ground truth and show competitive image and feature agreement. These results establish a practical hexagonal representation for quantitative CT.
\end{abstract}

\section{Introduction}
\label{sec:intro}
Image reconstruction from projection data is central to medical, biological and industrial imaging. Filtered backprojection offers efficient analytic reconstruction, while sparse-view and low-dose acquisitions motivate methods that explicitly incorporate the acquisition model and image constraints. Model-based iterative reconstruction (MBIR) places these ingredients in an optimization framework~\cite{Beister2012}. Its repeated forward and backprojection steps make the image representation important both for modeling the measurements and for computational cost.

A key design choice is how to represent the continuous attenuation map using finitely many coefficients. Cartesian pixels are simple and reproduce constant fields, but their piecewise-constant representation can introduce discretization artifacts~\cite{discrete1}. Overlapping bases offer a way to improve approximation without increasing the coefficient density. The projector must then integrate the chosen basis along rays and across the detector aperture. This connects representation design to the physical forward model: measured intensities are averaged before the logarithm used to form line-integral data.

Hexagonal sampling offers a principled geometric motivation for a different representation. For a disk-limited Fourier spectrum, an infinite triangular sampling lattice is the most sampling-efficient two-dimensional lattice: it needs $\sqrt{3}/2$ times the density of a square lattice to avoid spectral overlap~\cite{petersen1962sampling}. Its six equally distant nearest neighbors provide greater directional isotropy, while the larger alias-free circular bandwidth at equal density reduces susceptibility to aliasing under these sampling assumptions. On these centers, a compact three-direction linear tent reproduces affine functions and forms a partition of unity~\cite{condat2006threedirectional}. These properties motivate studying hexagonal bases at matched coefficient density, where lattice geometry and approximation order can be examined through constant and linear models.

The representation also matters for quantitative imaging. Radiomic intensity and texture descriptors depend on reconstruction and sampling choices~\cite{Reiazi2021CTRobustnessReview,Zwanenburg2020IBSI}; examining these descriptors complements pixelwise image comparisons. We therefore retain both image and radiomics analyses, using independent FORBILD simulation and acquired 2DeteCT measurements~\cite{2DeteCT}. Real measurements lack continuous analytic truth for a lattice-neutral comparison, so the paired low-dose/high-dose study uses a high-resolution Cartesian reconstruction as pseudo ground truth. We compare two coefficient densities at fixed image sampling and common solver budgets. This connects clean-data accuracy to noise sensitivity, particularly when coefficient resolution is limited.

This work contributes (i) explicitly normalized constant and linear hexagonal bases on common coefficient centers; (ii) their analytic ray footprints and effective detector-blur expressions for fan-beam CT, with matching backprojection; and (iii) a GPU implementation and matched Cartesian/hexagonal evaluation of reconstruction, runtime and radiomic features. The resulting model turns hexagonal sampling into a practical quantitative-reconstruction option: compact affine approximation, competitive projection cost and quantitative feature preservation at useful coefficient densities.

\section{Related work}
\subsection{Spline-based convolutional projection}
Alternative attenuation bases include compact, radially symmetric Kaiser--Bessel functions, whose approximation properties depend on their design and partition-of-unity behavior~\cite{unser}. Box splines provide a polynomial framework encompassing Cartesian and nonseparable representations. Entezari \emph{et al.} established closure under the Radon transform and explicit CT projection formulas~\cite{entezari2012box}. Zhang and Entezari subsequently expressed footprints and detector blur through directional convolutions, enabling efficient fan-beam implementations~\cite{zhang2019convolutionalforwardbackprojectionmodel,doi:10.1137/21M1464191}. These results supply the projection calculus used here.

\subsection{Hexagonal bases}
Hexagonal representations offer nonseparable bases: Voronoi-cell hex-splines~\cite{vandeville2004hexsplines} and three-direction box splines~\cite{condat2006threedirectional} provide established constructions with different supports. Here, the constant model is a Voronoi indicator and the linear model is a three-direction tent. We specialize the convolutional framework to this common-lattice pair, making their normalization, direction sets and fan-beam weights explicit. The comparison with Cartesian pixels examines reconstruction and quantitative features under one solver and execution policy.

\begin{figure}[t]
\centering
\includegraphics[width=\linewidth]{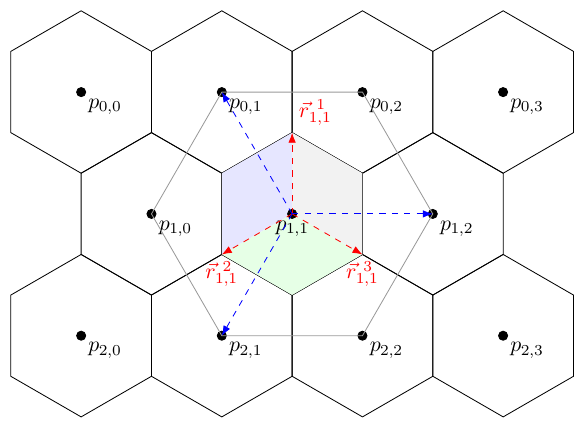}
\caption{Common hexagonal lattice. Red arrows are translated directions $\vec r^k$ forming H-Pixel's three rhombi; blue arrows are the neighbor directions in $\Xi^l$. H-Pixel occupies the shaded central cell; H-Linear's support is the thin gray outline through neighboring centers.}
\label{fig:bases}
\end{figure}

\section{Hexagonal basis functions}
\subsection{Geometry and normalization}
We call the constant and linear hexagonal models \emph{H-Pixel} and \emph{H-Linear} throughout; \emph{C-Pixel} denotes the Cartesian square-indicator model. SF denotes the additional separable-footprint baseline~\cite{long20103d}. Let $\Lambda_H=\{p_{i,j}\mid i,j\in\mathbb Z\}$ denote the coefficient centers in $\R^2$, and let $\vec x=(x_1,x_2)$ denote position. At each center, the translated displacement vectors are
\begin{equation}\label{eq:directions}
 \vec r^1=(0,d_t),\quad
 \vec r^2=(-\sqrt{3}d_t/2,-d_t/2),\quad
 \vec r^3=-\vec r^1-\vec r^2,
\end{equation}
with $D=|\det[\vec r^1,\vec r^2]|=\sqrt{3}d_t^2/2$. Define $\Xi^k=[\vec r^k,\vec r^{k+1}]$, with $\vec r^4=\vec r^1$, and $\Xi^l=[\vec r^2-\vec r^1,\vec r^3-\vec r^2,\vec r^1-\vec r^3]$. Any two neighbor columns have determinant magnitude $3D$.
For a two-dimensional direction matrix $\Xi$ with $m$ columns, its one-sided, unit-mass box spline is
$M_\Xi(\vec x)=\int_{[0,1]^m}\delta_2(\vec x-\Xi\vec t)\,d\vec t$.
The dimensionless scalar bases are
\begin{align}
 H^0(\vec x)&=D\sum_{k=1}^{3}M_{\Xi^k}(\vec x),\label{eq:h0}\\
 H^l(\vec x)&=3D M_{\Xi^l}(\vec x).
 \label{eq:hl}
\end{align}
The three rhombi in~\eqref{eq:h0} tile a pointy-top hexagon of width $\sqrt{3}d_t$, height $2d_t$ and area $3D$ (Fig.~\ref{fig:bases}). Thus H-Pixel is its indicator. H-Linear is a unit-peak tent supported by the six neighboring centers, forming a flat-top hexagon of radius $\sqrt{3}d_t$. Its integral is also $3D$. With $\vec x=a(\vec r^2-\vec r^1)+b(\vec r^3-\vec r^2)$, it is
\begin{equation}\label{eq:hat}
 H^l(\vec x)=\pos{1-\max(|a|,|b|,|a-b|)}.
\end{equation}
For either model $\alpha\in\{0,l\}$, define $H_{i,j}^{\alpha}(\vec x)=H^{\alpha}(\vec x-p_{i,j})$ and $f(\vec x)=\sum_{i,j}V_{i,j}H_{i,j}^{\alpha}(\vec x)$, where $V_{i,j}$ has attenuation units.

Both models use the same center lattice: horizontal spacing $\sqrt{3}d_t$, row spacing $3d_t/2$ and a half-column shift on odd rows. Both form partitions of unity on the untruncated lattice, with cell area $3D$; H-Linear also reproduces affine functions. For Cartesian spacing $d$, we choose $3D=d^2$. The two hexagonal models then have identical centers, masks and finite coefficient counts. Boundary values use zero extension. H-Linear has support area $9D=3d^2$. Its three direction factors yield piecewise-quadratic ray footprints. Appending detector blur raises the degree by one for nondegenerate widths.

\subsection{Footprints and detector blur}
For a source $\vec s$, a unit ray direction $\vec d(u)$, a generic dimensionless basis $\phi$, and detector bin $[u_n^-,u_n^+]$ of width $\Delta u$, the linear model is
\begin{equation}\label{eq:measurement}
 A_{n,(i,j)}=\frac1{\Delta u}\int_{u_n^-}^{u_n^+}\int_{\R}
 \phi(\vec s+t\vec d(u)-p_{i,j})\,dt\,du.
\end{equation}
Here $n$ indexes a view/bin and $(i,j)$ a coefficient center. Each basis contributes a \emph{column}; flattening $V_{i,j}$ gives $x$ in $Ax=b$. Photon data average intensities before taking a negative logarithm, generally differing from this line-average model.

For a row of signed directions $\vec w=[w_1,\ldots,w_m]$, the one-dimensional box spline is $M_{\vec w}(q)=\int_{[0,1]^m}\delta_1(q-\sum_k w_k t_k)\,d\vec t$. This is the unit-mass convolution of signed uniform intervals. For $m\ge2$ nonzero widths,
\begin{equation}\label{eq:finite}
 M_{\vec w}(q)=\frac{\displaystyle\sum_{\vec\epsilon\in\{0,1\}^m}
 (-1)^{|\vec\epsilon|}\pos{q-\vec\epsilon\cdot\vec w}^{m-1}}
 {(m-1)!\prod_k w_k}.
\end{equation}
Here $|\vec\epsilon|=\sum_k\epsilon_k$. Zero widths are omitted as delta distributions; a single remaining width gives its normalized interval indicator. Lower-order or piecewise limits handle degenerate directions.

At detector coordinate $u$, let $\vec n=\vec n(u)$ be the unit normal to $\vec d(u)$, oriented toward increasing detector coordinate. Define $(P_u\phi)(q)=\int_{\R}\phi(q\vec n+t\vec d(u))\,dt$, where $q$ is physical signed distance along $\vec n$, distinct from detector coordinate $u$. Projection commutes with directional convolution~\cite{entezari2012box}, giving
\begin{equation}\label{eq:footprints}
 \begin{aligned}
 (P_u H^0)(q)&=D\sum_{k=1}^3 M_{\vec n^T\Xi^k}(q),\\
 (P_u H^l)(q)&=3D M_{\vec n^T\Xi^l}(q).
 \end{aligned}
\end{equation}
The one-sided convention retains each rhombus translation in H-Pixel.

For bin center $\vec z$, edges $\vec z_\pm$, and $\vec v=\vec z-\vec s$, intersect edge rays with the line through $p_{i,j}$ normal to $\vec v$:
\begin{equation}
 \vec e_\pm=\vec s+
 \frac{(p_{i,j}-\vec s)\cdot\vec v}
 {(\vec z_\pm-\vec s)\cdot\vec v}(\vec z_\pm-\vec s).
\end{equation}
Using $\vec n$ at the bin center, set
$q_-=\vec n\cdot(\vec e_--p_{i,j})$,
$\omega=\vec n\cdot(\vec e_+-\vec e_-)>0$. Appending the signed direction $-\omega$ gives the effective blurred weights
\begin{align}\label{eq:blurred}
 \widetilde A_{n,(i,j)}^{0}&=D\sum_{k=1}^3 M_{[\vec n^T\Xi^k,-\omega]}(q_-),\\
 \widetilde A_{n,(i,j)}^{l}&=3D M_{[\vec n^T\Xi^l,-\omega]}(q_-).\nonumber
\end{align}
The unit-mass appended interval averages locally parallel ray footprints across $[q_-,q_-+\omega]$. It is exact for parallel rays and an effective-blur approximation in fan-beam geometry~\cite{zhang2019convolutionalforwardbackprojectionmodel}. Below, $A$ denotes each implemented projector; the hexagonal models use $\widetilde A$. Backprojection uses the transpose of the same weights, without an empirical scale factor.

\begin{figure*}[t]
\centering
\includegraphics[width=\textwidth]{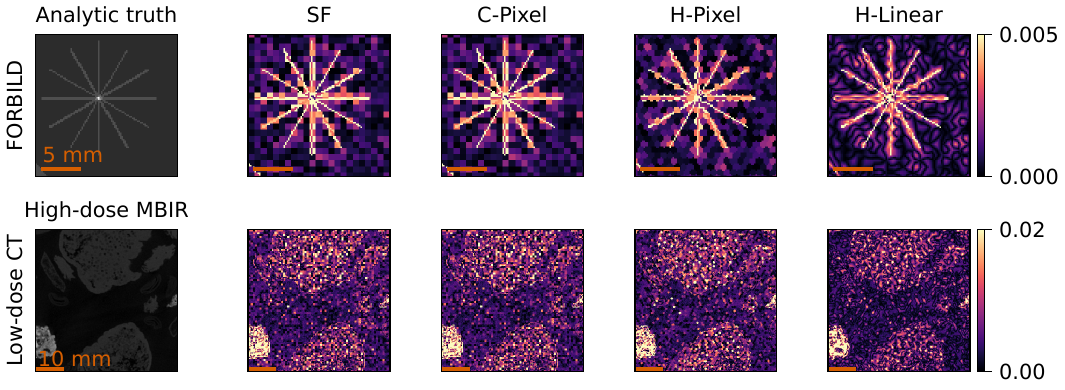}
\caption{Common physical regions and error scales for all models. Top: FORBILD truth and absolute errors, $n=128$, 300 iterations, seed 2032; truth window [0,0.06]~mm$^{-1}$, 5-mm bars. Bottom: high-dose MBIR reference and low-dose deviations, 2DeteCT slice 525, $n=256$, 100 iterations; reference window [0,0.12]~mm$^{-1}$, 10-mm bars. Error scales saturate at the shown maxima (mm$^{-1}$).}
\label{fig:images}
\end{figure*}

\section{Experiments and discussion}
\label{sec:results}
\subsection{Study setup}
We compare SF, C-Pixel, H-Pixel and H-Linear on FORBILD and 2DeteCT. H-Pixel/C-Pixel compares constant bases; H-Linear/H-Pixel tests affine approximation on common centers. All methods solve nonnegative least squares from zero using FISTA~\cite{beck2009fista}, power-estimated step sizes and majorant backtracking, without an explicit penalty. Every model and resolution uses a fixed budget: 300 iterations for FORBILD and 100 for 2DeteCT. References do not select iterations. We vary coefficient density while holding measurements, evaluation points and feature settings fixed.

\textbf{Simulated data.} Independent analytic clipped-ellipse integrals and 128-point detector quadrature generate $180\times384$ FORBILD measurements. The study includes clean line averages, clean intensity averages and five paired Poisson realizations ($N_0=20{,}000$, seeds 2032--2036). Cartesian $n^2$ coefficient fields with $n=128,256$ cover 110~mm, with disk-masked centers and equal-density hexagonal grids. Generation is independent of every reconstruction matrix.

\textbf{Real-world CT data.} Laboratory 2DeteCT supplies $180\times956$ measurements for slices 525, 725 and 925, reserved after protocol development~\cite{2DeteCT}. Both low-dose Mode~1 and high-dose Mode~2 reconstructions are scored against the same-slice high-dose, $1024^2$ Cartesian MBIR reference from the dataset authors' all-view ASTRA pipeline. This is pseudo ground truth, not an independent attenuation map. Mode~1 has approximately 1/30 the exposure at the same voltage/filtration. Cartesian $n=256,512$ fields span 233.06~mm, twice the scored width. They therefore contain about 128/256 coefficient intervals across the crop, bridging FORBILD's two resolutions: spacings are 0.910/0.455~mm versus 0.859/0.430~mm. Hexagonal cell areas match at each level; no fitted registration or intensity scaling is used.

\textbf{Evaluation.} Native synthesis uses fixed $512^2$ physical grids and circular masks. For FORBILD and 2DeteCT, PyRadiomics~\cite{10.1158/0008-5472.CAN-17-0339} extracts 18 first-order, 24 GLCM and 16 GLRLM features. Ten image-wise bins retain the original protocol; twenty bins and two frozen physical bin widths test sensitivity. For feature $f$ and reference $f_r$, deviation is $200|f-f_r|/(|f|+|f_r|)$ percent, zero for two zeros. We average each feature's error across cases, then take its family median. The supplement specifies sampling and extraction; complete per-feature measurements and image metrics remain in the reproduction archive.

\subsection{Projector implementation and speed}
\textbf{Implementation.} Equation~\eqref{eq:finite} defines the weights. The GPU code handles signed and degenerate directions and integrates short blur intervals on normalized spline-knot intervals to limit cancellation. Each thread processes a coefficient/view pair over its support bins; cached geometry and reusable buffers serve forward and transpose calls with identical weights. C-Pixel, H-Pixel and H-Linear share this execution policy, float64 preparation/accumulation and float32 cache/footprint arithmetic. The supplement summarizes numerical evaluation and validation.

Independent ray/bin integration and adjoint probes pass in 36 GPU geometry configurations; worst footprint error is 0.102\%. All 176 independent-reference primitive probes for the three basis projectors pass the $2\times10^{-5}$ scaled-absolute-error gate. These test the implementation in specified domains, not all geometries.
\begin{table}[t]
\centering
\setlength{\tabcolsep}{4pt}
\caption{Independent median forward/backward pair times (ms): 33 synchronized samples per model/geometry. Initialization is separate.}
\label{tab:independent-timing}
\begin{tabular}{lrrr}
\toprule
Model & FORBILD & 180 views & 450 views \\
\midrule
SF & 2.75 & 17.89 & 84.72 \\
C-Pixel & 1.61 & 18.89 & 99.86 \\
H-Pixel & 2.46 & 27.00 & 146.90 \\
H-Linear & 2.32 & 26.94 & 156.54 \\
\bottomrule
\end{tabular}
\end{table}

Table~\ref{tab:independent-timing} summarizes 33 synchronized pairs in three fresh processes per model/geometry on an RTX 3090 (CUDA 12.3), including host allocation and transfers. Cartesian grid widths are 128, 512 and 1024, respectively. H-Linear/H-Pixel ratios are 0.94, 1.00 and 1.07. C-Pixel leads the three shared-engine basis projectors; SF leads the acquired-data timings. H-Linear provides affine approximation at near H-Pixel cost, with runtime also affected by bin overlap, support orientation and memory access.

\subsection{Approximation order and noise sensitivity}
\begin{table}[t]
\centering
\setlength{\tabcolsep}{3pt}
\caption{FORBILD PSNR (dB), 300 iterations for all methods, fixed $512^2$ evaluation. Clean: intensity-average data; noisy: five paired Poisson seeds. $n=128/256$ Cartesian coefficient widths have equal-density hexagonal counterparts.}
\label{tab:resolution-simulation}
\begin{tabular}{lrrrr}
\toprule
 & \multicolumn{2}{c}{$n=128$} & \multicolumn{2}{c}{$n=256$} \\
\cmidrule(lr){2-3}\cmidrule(lr){4-5}
Model & Clean & Noisy & Clean & Noisy \\
\midrule
SF & 26.91 & 26.40 & 27.89 & 21.71 \\
C-Pixel & 26.91 & 26.40 & 27.89 & 21.71 \\
H-Pixel & 26.97 & 26.46 & 27.97 & 21.80 \\
H-Linear & 28.21 & 27.66 & 30.52 & 24.64 \\
\bottomrule
\end{tabular}
\end{table}

Table~\ref{tab:resolution-simulation} supports the central approximation argument: H-Linear improves on both pixel models at both densities, for clean and noisy data. H-Linear achieves this on H-Pixel's existing centers with only three directional factors. At the coarser $n=128$ grid, its noisy PSNR is 27.66~dB, compared with 26.40~dB for SF/C-Pixel and 26.46~dB for H-Pixel. Figure~\ref{fig:images} compares all methods in the same physical regions.

Resolution also determines how this approximation is used. Doubling the coefficient count in each direction improves clean-data PSNR but reduces noisy PSNR for the unpenalized iterative methods. H-Linear's best noisy result occurs at the coarser grid. Its compact affine representation is therefore particularly useful when a coarse coefficient grid is needed to limit noise sensitivity. The coarse-grid comparison demonstrates improved reconstruction without increasing coefficient density, supporting the practical sampling motivation. These fixed-budget results concern reconstruction accuracy; they do not separately measure aliasing or equal convergence.

\subsection{Resolution and quantitative CT features}
\label{sec:resolution}
\begin{table*}[t]
\centering
\setlength{\tabcolsep}{3pt}
\caption{2DeteCT feature deviations (\%), 10 bins, three slices, 100 iterations for all methods. $n$ is the Cartesian coefficient-field width; hexagonal grids match cell area. FO/CM/RL are first-order/GLCM/GLRLM family medians. Both doses use the same high-dose MBIR reference; lower is closer agreement.}
\label{tab:resolution-real-count10}
\begin{tabular}{lrrrrrrrrrrrr}
\toprule
Model & \multicolumn{3}{c}{Low dose; $n=256$} & \multicolumn{3}{c}{Low dose; $n=512$} & \multicolumn{3}{c}{High dose; $n=256$} & \multicolumn{3}{c}{High dose; $n=512$} \\
\cmidrule(lr){2-4}\cmidrule(lr){5-7}\cmidrule(lr){8-10}\cmidrule(lr){11-13}
 & FO & CM & RL & FO & CM & RL & FO & CM & RL & FO & CM & RL \\
\midrule
SF & 31.44 & 18.23 & 62.84 & 45.91 & 25.02 & 68.81 & 2.27 & 9.56 & 24.09 & 2.84 & 8.30 & 31.56 \\
C-Pixel & 31.44 & 18.23 & 62.83 & 45.91 & 25.02 & 68.81 & 2.27 & 9.56 & 24.09 & 2.84 & 8.30 & 31.55 \\
H-Pixel & 32.08 & 17.40 & 50.63 & 47.19 & 24.47 & 58.28 & 2.26 & 8.47 & 24.26 & 3.02 & 8.91 & 31.29 \\
H-Linear & 28.42 & 12.65 & 32.05 & 33.82 & 17.81 & 56.08 & 2.44 & 6.52 & 18.73 & 1.83 & 4.27 & 18.70 \\
\bottomrule
\end{tabular}
\end{table*}

Table~\ref{tab:resolution-real-count10} tests this connection on acquired data with a fixed physical texture scale. At the coarser coefficient grid, H-Linear reduces low-dose first-order, GLCM and GLRLM deviations relative to C-Pixel from 31.44/18.23/62.83\% to 28.42/12.65/32.05\%. H-Pixel also improves both texture-family summaries over C-Pixel. H-Linear improves all three low-dose families over H-Pixel at both densities. The representation benefit therefore extends beyond pixelwise image scores to quantitative texture.

The high-dose comparison reinforces the value of the compact tent at the coarser density. H-Linear lowers GLCM/GLRLM deviations relative to C-Pixel from 9.56/24.09\% to 6.52/18.73\%, while first-order deviation is comparable (2.44\% versus 2.27\%). Without analytic real-world truth, these results establish competitive agreement with the high-dose Cartesian pseudo-reference, rather than a lattice-neutral accuracy ranking. They support hexagonal affine approximation for quantitative reconstruction without requiring a finer coefficient grid.

The two densities also delimit that benefit. At the finer grid, H-Linear's ten-bin low-dose GLRLM deviation is only slightly below H-Pixel's (56.08\% versus 58.28\%). The twenty-bin and fixed-width sensitivities in the supplement show that feature agreement depends on discretization as well as coefficient density. These results favor choosing coefficient density and feature scale together. Both image and feature agreement remain relative to Cartesian MBIR references; the three-slice study does not establish independent attenuation accuracy or clinical repeatability.

\section{Conclusion}
We developed constant and linear hexagonal projectors on common centers, using box spline calculus to derive analytic footprints and effective fan-beam blur, with efficient GPU forward and backprojection. The compact affine tent improves FORBILD accuracy over pixels and gives its best noisy result at coarser coefficient density. Real-world reconstructions achieve competitive agreement with high-dose, high-resolution Cartesian pseudo ground truth. These findings support practical hexagonal reconstruction motivated by sampling efficiency; independent physical-reference validation remains future work.

\noindent\textbf{Acknowledgments.}
OpenAI Codex assisted with formating and grammer checking for the manuscript, and preparation and formatting of result plots according to controlling prompts.

\clearpage
\nobalance
\bibliographystyle{IEEEbib}
\bibliography{strings,refs}
\end{document}